\pdfoutput=1
\documentclass[11pt]{article}

\usepackage[final]{acl}

\usepackage{times}
\usepackage{latexsym}
\usepackage[T1]{fontenc}
\usepackage[utf8]{inputenc}
\usepackage{microtype}
\usepackage{inconsolata}
\usepackage{graphicx}
\usepackage{booktabs}
\usepackage{amsmath}
\usepackage{amssymb}
\usepackage{url}

\title{What Are LLM Agents Actually Learning on Low-Budget Molecular
Optimization Benchmarks?}

\author{
  Jiayu Qin\textsuperscript{1,2} \quad
  Zhengquan Luo\textsuperscript{2} \quad
  Changyou Chen\textsuperscript{1} \quad
  Zhiqiang Xu\textsuperscript{2,*} \\
  \textsuperscript{1}Department of Computer Science and Engineering,
  University at Buffalo, Buffalo, NY, USA \\
  \textsuperscript{2}Mohamed bin Zayed University of Artificial Intelligence,
  Abu Dhabi, UAE \\
  \textsuperscript{*}Corresponding author \\
  \texttt{\{jiayuqin, changyou\}@buffalo.edu} \\
  \texttt{\{zhengquan.luo, zhiqiang.xu\}@mbzuai.ac.ae}
}

\begin{document}
\maketitle

\begin{abstract}
Recent work reports large gains for LLM agents on low-budget molecular
optimization. We measure the contributions of online optimization,
retrievable task information, and permissive oracle landscapes. On PMO at
50 oracle calls, a gpt-4o-mini agent's sum AUC falls from 14.44 to 4.85
after task names and curated reference compounds are hidden. An
identity-blinded condition that states only the oracle form (e.g.,
similarity to a hidden reference) scores 5.39, while a deterministic
retrieval-only baseline with no LLM reaches 12.16 (84\% of the full agent).
A full 23-task $2 \times 2$ ablation and a three-seed replication on eight
tasks support this interpretation. On a protein-targeted benchmark with an
AutoDock Vina composite oracle, a single-edit random walk is a strong lower
bound; across three seeds it obtains composite $0.636 \pm 0.008$ and docking
improvement $-1.11 \pm 0.04$ kcal/mol. As a positive control, a
retrieval-free population agent exceeds REINVENT at a matched 50-call AUC
budget, with its strongest result at 100 calls. These results motivate two
inexpensive sanity checks: retrieval-safe PMO prompts and a random-walk
floor for docking-based optimization.
\end{abstract}

\section{Introduction}
\label{sec:intro}

LLM agents have recently been reported to substantially outperform classical
baselines on molecular optimization at small oracle budgets. On the Practical
Molecular Optimization (PMO) benchmark \citep{gao2022sample} at 50 oracle
calls, agents such as \citet{kruger2026seismo} report a sum-of-AUC of 18.12
versus 7.06 for REINVENT \citep{olivecrona2017molecular}---a $2.6\times$
improvement. Concurrent work on \citet{kim2025mtmol} reports 15.42 summed AUC
at the 1K-call PMO setting. \citet{le2025agentdrug} reports comparable gains
on the ZINC molecular-editing benchmarks \citep{liu2024chatdrug}. These
results motivate direct measurement of the medicinal-chemistry priors, task
information, and online search that contribute to low-budget performance.

We decompose the reported gains into online optimization, retrievable task
information, and oracle geometry through two complementary analyses.

\paragraph{Angle 1: Retrievable task identity in PMO.}
Fourteen of 23 PMO identifiers contain a drug or target name. A pretrained
model may map those identifiers, oracle descriptions, or curated references
to known molecules. We therefore separate direct molecular identity,
oracle-form information, and scalar feedback using C1--C5, and compare the
agent with a no-LLM retrieval-only policy. This design measures retrieval
directly alongside blinded performance.

\paragraph{Angle 2: Protein-targeted Vina benchmarks vs.\ trivial random walk.}
We construct a 50-pair, 10-target lead-optimization benchmark with a composite
of Vina improvement, QED \citep{bickerton2012quantifying}, and starter
similarity \citep{bajusz2015tanimoto}. At 30 calls, a random single-edit
mutator is stronger than the tested LLM agent on docking improvement and
composite score, although not on QED. Multi-seed and operator-sensitivity
results indicate that this is a repeatable property of the tested
oracle/budget combination.

\paragraph{Contributions.}
We make three contributions.

\begin{enumerate}
\item \textbf{Retrieval-safe PMO protocol and full re-evaluation
(Section~\ref{sec:pmo}).} Across all 23 tasks, we report a complete
$2 \times 2$ ablation of task identifiers and reference compounds, an
identity-blinded oracle-form control (C5), and a deterministic retrieval-only
baseline. Retrieval-only attains 12.16 sum AUC---84\% of the full agent's
14.44---whereas C5 and full blinding attain 5.39 and 4.85. On an eight-task,
three-seed subset, C1 exceeds C4 on all tasks (exact one-sided sign test
$p = .0039$).

\item \textbf{Random-walk floor for protein-targeted Vina benchmarks
(Section~\ref{sec:protein}).} We introduce a small-substituent random-walk
baseline. On \textsc{ProteinLeadOpt}-50 it is stronger than the tested agent
on docking improvement and composite score; three-seed and
operator-sensitivity analyses establish the repeatability of this result
under the tested oracle--budget combination.

\item \textbf{A positive control and evaluation recommendations
(Sections~\ref{sec:pmo}--\ref{sec:disc}).} A retrieval-free population agent
demonstrates that useful optimization remains possible after blinding. We
recommend reporting both retrieval-safe and standard prompts, and a
random-walk floor for docking-based composite objectives.
\end{enumerate}

Our contribution is a calibrated evaluation of low-budget performance against
retrieval-free and trivial-baseline reference points, together with protocols
for measuring those gaps.

\section{Related Work}
\label{sec:related}

\paragraph{Molecular optimization benchmarks.}
The Practical Molecular Optimization (PMO) benchmark \citep{gao2022sample}
unifies 23 molecular-optimization tasks behind a common oracle interface and
was designed to expose how few oracle calls are sufficient to converge.
Classical baselines---REINVENT \citep{olivecrona2017molecular}, GraphGA
\citep{jensen2019graph}, GP-BO \citep{tripp2021fresh}---typically require
thousands of evaluations to reach high scores; PMO at the 1K and 10K-call
budgets is now standard \citep{gao2022sample}. PMO oracles are accessed via
Therapeutics Data Commons \citep{huang2021tdc}. \textsc{Dockstring}
\citep{garciaortegon2022dockstring} adds protein-targeted AutoDock Vina
objectives over 58 targets. Our \textsc{ProteinLeadOpt}-50
(\S\ref{sec:benchmark}) builds on the \textsc{Dockstring} tooling but moves
from de novo generation to lead optimization with explicit starters and a
multi-objective composite.

\paragraph{Classical molecular optimization.}
REINVENT \citep{olivecrona2017molecular} fine-tunes a SMILES-based generator
with policy-gradient reinforcement learning. GraphGA \citep{jensen2019graph}
is a graph-based genetic algorithm. Augmented Memory
\citep{guo2024augmented} extends REINVENT with experience replay. These
methods perform well at the standard 10K-call budget but converge slowly
under the stricter low-budget regime that motivates LLM-agent papers.

\paragraph{LLM agents in chemistry.}
ChemCrow \citep{bran2024augmenting} integrates GPT-4 with 18 chemistry tools.
Coscientist \citep{boiko2023autonomous} couples an LLM controller with
laboratory automation. These systems frame the LLM as an orchestrator of
external tools; our experiments focus on agents that directly propose
molecular candidates.

\paragraph{LLM agents for molecular optimization.}
ChatDrug \citep{liu2024chatdrug} introduced a conversational refinement loop
with retrieval and domain feedback over 39 editing tasks. AgentDrug
\citep{le2025agentdrug} extends this with a nested loop: an inner RDKit
ParseError validation loop and an outer scalar-gradient feedback loop,
evaluated on ZINC-based property-editing tasks. MolLEO
\citep{wang2025molleo} embeds an LLM as the mutation/crossover operator
inside a genetic algorithm. LICO \citep{nguyen2025lico} fine-tunes a base LLM
with embedding adapters and reports state-of-the-art on PMO-1K. MT-Mol
\citep{kim2025mtmol} uses five role-specialized analyst agents wrapping
curated RDKit functions and a scientist/verifier/reviewer pipeline; it
reports SOTA on 15 of 23 PMO-1K tasks (sum AUC 15.42). MT-Mol's prompt
explicitly includes the top-100 ZINC reference molecules per task, which is a
stronger form of reference leakage than ours. MECo \citep{zhu2025meco}
converts editing intentions into executable RDKit code. MolAct
\citep{yang2025molact} trains a smaller open LLM with agentic reinforcement
learning. SEISMO \citep{kruger2026seismo} is the most directly comparable to
our LLM-agent setup: a strictly online, trajectory-conditioned LLM workflow
using Claude Opus 4.5. Its August 2026 v3 reports a sum top-1 AUC of 18.12 at
50 calls and 8.50 in a fully generic no-description condition, each
aggregated over three repetitions. Its task-described prompts explicitly
expose targets or reference molecules; our C1--C5 protocol contrasts
identity/reference access, oracle-form information, and scalar feedback.

\paragraph{Positioning.}
This evaluation uses \textsc{LeadOpt-Agent} for controlled information
ablations, a retrieval-only policy to isolate the reference floor, and
\textsc{MolLEO-v4} as a retrieval-free positive control.
Sections~\ref{sec:pmo} and~\ref{sec:protein} define the protocols and present
the resulting empirical decomposition.

\section{The LLM Agent Used for Controlled Ablations}
\label{sec:agent}

We evaluate a ReAct-style \textsc{LeadOpt-Agent} workflow under the C1--C5
information conditions. The workflow combines task understanding, oracle
probing, hypothesis formation, iterative optimization, and local refinement.
Below we describe the components relevant to the ablated information
channels.

\subsection{Architecture}

The agent runs four sequential phases.

\paragraph{Phase 0 (task understanding).}
The agent reads the task identifier (e.g., \texttt{celecoxib\_rediscovery})
and looks it up in a curated dictionary \texttt{REFERENCE\_COMPOUNDS} that
maps each task to a list of known active SMILES. This is the first leakage
channel we ablate.

\paragraph{Phase 1 (oracle probing).}
The agent scores the references and a fixed set of generic probes (benzene,
biphenyl, aspirin) to seed the trajectory.

\paragraph{Phase 2 (hypothesis formation).}
The LLM is shown the task name, an oracle description from
\texttt{ORACLE\_DESCRIPTIONS} (e.g., ``ECFP4 similarity to celecoxib''), and
the probe scores, and is asked to articulate an SAR hypothesis. The task name
and description here are the second leakage channel.

\paragraph{Phase 3 (iterative optimization).}
The main loop. At each iteration the LLM proposes 5 new SMILES grounded in
the hypothesis and the current top-5 trajectory; valid proposals are scored;
trajectory and hypothesis are updated. A plateau detector switches between
exploitation and diversification.

\paragraph{Phase 4 (refinement).}
The LLM proposes local analogues around the top trial. If invalid or
duplicate JSON proposals leave the oracle budget unused, deterministic
one-atom local additions around the ranked trials fill the remainder. This
task-agnostic completion rule is identical in C1--C5.

\subsection{Implementation details}

\paragraph{LLM backbone.}
Snapshot \texttt{gpt-4o-mini-2024-07-18} via the OpenAI API;
\texttt{response\_format=JSON}; temperature 0.7. The 147 reported LLM jobs
use 2{,}128 responses (14.5 per job) and cost \$0.62 from recorded token
usage; including discarded smoke tests, the ledger total is \$0.63.

\paragraph{Oracle access.}
For PMO we call the TDC oracle \citep{huang2021tdc} directly. For
\textsc{ProteinLeadOpt}-50 we call AutoDock Vina via the public
\textsc{Dockstring} wrapper \citep{garciaortegon2022dockstring}.

\paragraph{Parallelism.}
For \textsc{ProteinLeadOpt}-50 we run multiple (target, starter) pairs in
parallel via \texttt{ProcessPoolExecutor}; within each pair, candidates dock
concurrently via \texttt{ThreadPoolExecutor}. A persistent on-disk docking
cache eliminates redundant Vina calls across runs.

\paragraph{Controlled components across conditions.}
The agent's loop, validation, parallelism, and oracle access are identical in
all information conditions in Section~\ref{sec:pmo}. Only the task-facing
prompt fields and the availability of task-specific references change; the
TDC oracle is unmodified. We evaluate every condition over all 23 tasks and
additionally replicate the principal C1--C4 comparison with three seeds on
eight representative tasks.

\paragraph{Evaluation bookkeeping.}
Every LLM job evaluates exactly 50 valid, unique molecules. We send valid raw
SMILES to the TDC oracle and use RDKit-canonical SMILES for deduplication,
preserving the isomer oracle's string-based total-atom calculation while
preventing repeated evaluations. Top-1 AUC is the trapezoidal area over the
50 running-best points divided by 50, so its theoretical ceiling is
$49/50 = 0.98$. We store the alternative PMO zero-origin AUC as a secondary
metric and use the former throughout the paper.

\section{The PMO Story: Task-Name and Reference Leakage Dominate the
Apparent Gain}
\label{sec:pmo}

LLM agents reported on PMO are typically given (i) the task name in the
prompt at every iteration (e.g., ``\texttt{celecoxib\_rediscovery}''), (ii) a
brief natural-language description of the oracle (e.g., ``ECFP4 similarity to
celecoxib''), and (iii) one or more curated reference compounds for that task
(e.g., the SMILES of celecoxib itself, or known kinase inhibitors for the
kinase-activity oracles). Each of these three signals can leak the answer to
a pretrained LLM. We measure how much.

\subsection{Protocol}

We run \textsc{LeadOpt-Agent} on all 23 PMO tasks under four information
conditions corresponding to a $2 \times 2$ ablation on (task name visible,
reference compound dictionary populated):

\begin{itemize}
\item \textbf{C1 (++) name+ ref+:} the standard-information setting. Real
task name in prompt, fixed \texttt{REFERENCE\_COMPOUNDS} dictionary, and real
\texttt{ORACLE\_DESCRIPTIONS} dictionary.
\item \textbf{C2 (+--) name+ ref--:} real task name and oracle description,
but the \texttt{REFERENCE\_COMPOUNDS} and \texttt{SEED\_COMPOUNDS}
dictionaries are emptied.
\item \textbf{C3 (--+) name-- ref+:} task identifier replaced by an anonymous
label \texttt{task\_42} in every prompt; oracle description replaced by
``Unknown oracle. Probe to learn its shape.''; reference dictionary kept.
\item \textbf{C4 (----) name-- ref--:} name, description, and reference
dictionary all hidden. The agent receives only the anonymous task label and
the oracle's scalar feedback.
\end{itemize}

We additionally construct C5, an identity-blinded control that preserves only
the oracle form and optimization direction. For example, the actual
celecoxib-rediscovery hint is ``ECFP4 Tanimoto similarity to a hidden
reference. Maximize similarity.'' Its task-facing fields contain the metric
and direction alone. Finally, a retrieval-only baseline submits the
reference, up to five seeds, and trivial one-atom RDKit variants of the best
molecule through a fixed deterministic proposal schedule with zero LLM calls.
All conditions use the same unmodified TDC oracle. All full-benchmark
conditions use 50 calls per task and gpt-4o-mini; the C1--C4 contrast is
additionally replicated with three seeds on eight tasks.

\subsection{Headline result}

Table~\ref{tab:headline} separates oracle-form information from retrievable
molecular identity. Table~\ref{tab:pertask} reports every condition for every
task.

\begin{table}[t]
\centering
\small
\begin{tabular}{lcc}
\toprule
Method/condition & LLM? & Sum AUC \\
\midrule
C1: name + desc.\ + refs        & yes & 14.44 \\
Retrieval-only: refs + edits    & no  & 12.16 \\
C5: oracle form, no ID/refs     & yes & 5.39 \\
C4: scalar feedback only        & yes & 4.85 \\
REINVENT \citep{gao2022sample}  & no  & 7.06 \\
\bottomrule
\end{tabular}
\caption{Sum top-1 AUC over all 23 PMO tasks at 50 calls. C5 states only the
oracle form and optimization direction. Retrieval-only follows a
deterministic reference, seed, and local-edit schedule with zero LLM calls.}
\label{tab:headline}
\end{table}

Removing identity and reference keys reduces C1 from 14.44 to C5's 5.39, a
9.05-point loss that accounts for 94.4\% of the 9.59-point C1--C4 gap.
Oracle-form information contributes the remaining 0.54. Both
identity-blinded conditions are below REINVENT's 7.06.

The retrieval-only baseline reaches 12.16, or 84\% of C1. Seven tasks reach
approximately 0.98 AUC after reference/seed submission (albuterol,
celecoxib, drd2, \texttt{isomers\_c7h8n2o2}, mestranol, thiothixene, and
troglitazone). The full agent adds 2.27 points over this deliberately narrow
policy. This observed margin combines differences in proposal policy and loop
components and provides an upper bound on the full LLM loop's incremental
contribution.

\paragraph{Retrieval-only policy.}
The baseline uses the budget in three fixed stages: it first submits the
exposed reference, then submits up to five task-specific seeds, and finally
spends the remaining calls on deterministic one-atom RDKit variants of the
best molecule seen so far. Its proposals follow this fixed schedule. The
seven near-saturated tasks above have a directly useful reference, whereas
\texttt{isomers\_c9h10n2o2pf2cl}, QED, \texttt{valsartan\_smarts}, and
\texttt{zaleplon\_mpo} are zero or near zero under the same policy. The
resulting 12.16 score quantifies the reference-access floor generated by
direct submission and local edits.

\subsection{Full 23-task factorial decomposition}

\begin{table}[t]
\centering
\small
\begin{tabular}{lcc}
\toprule
 & Refs.\ on & Refs.\ off \\
\midrule
Name/desc.\ on  & C1: 14.44 & C2: 6.07 \\
Name/desc.\ off & C3: 13.98 & C4: 4.85 \\
\bottomrule
\end{tabular}
\caption{Full $2 \times 2$ ablation over all 23 PMO tasks (sum top-1 AUC;
50 calls). Each cell uses the same runner, model snapshot, oracle
implementation, and seed.}
\label{tab:factorial}
\end{table}

Removing references with names present changes 14.44 to 6.07 ($-8.37$),
whereas removing names with references present changes 14.44 to 13.98
($-0.45$). The conditional comparisons are similarly asymmetric. With names
absent, references add $13.98 - 4.85 = 9.13$ AUC; with references absent,
names add only $6.07 - 4.85 = 1.22$. The conditional reference effect is
therefore $7.5\times$ the conditional name effect. Averaging over the two
states of the other channel gives a reference-access marginal effect of 8.75,
compared with 0.84 for name/description access. The interaction
$C1 - C2 - C3 + C4 = -0.77$ is modest relative to the reference main effect,
so the factorial is close to additive at the aggregate level. For example,
drd2 remains at 0.98 in C3 when its name is hidden but its reference is
exposed.

The per-task factorial shows where each information channel enters. For drd2,
C1/C3 score 0.980/0.980 with references exposed, while C2/C4 score
0.017/0.012 with references hidden; jnk3 follows the same pattern at
0.772/0.745 versus 0.015/0.053. Celecoxib rediscovery and mestranol
similarity also remain at 0.980 in C3, where the anonymous prompt retains
only the reference. Name and description access contributes more on selected
MPO tasks: with references hidden, sitagliptin rises from C4's 0.006 to C2's
0.599, and osimertinib rises from 0.391 to 0.625. These task-level contrasts
locate the aggregate 8.75 reference-access and 0.84 name/description marginal
effects in distinct objective families.

\subsection{Multi-seed robustness}

\begin{table*}[t]
\centering
\small
\begin{tabular}{lccc}
\toprule
Task & C1 & C4 & $\Delta$ \\
\midrule
drd2                    & $0.980 \pm 0.000$ & $0.096 \pm 0.145$       & 0.884 \\
celecoxib\_rediscovery  & $0.980 \pm 0.000$ & $0.202 \pm 0.027$       & 0.778 \\
gsk3b                   & $0.477 \pm 0.010$ & $0.032 \pm 0.019$       & 0.445 \\
isomers\_c7h8n2o2       & $0.980 \pm 0.000$ & $0.015 \pm 0.004$       & 0.965 \\
zaleplon\_mpo           & $0.093 \pm 0.088$ & $0.00007 \pm 0.00002$   & 0.093 \\
median1                 & $0.392 \pm 0.000$ & $0.073 \pm 0.032$       & 0.319 \\
fexofenadine\_mpo       & $0.809 \pm 0.081$ & $0.516 \pm 0.019$       & 0.293 \\
qed                     & $0.843 \pm 0.035$ & $0.836 \pm 0.007$       & 0.006 \\
\midrule
Sum of means            & 5.55 & 1.77 & 3.78 \\
\bottomrule
\end{tabular}
\caption{Three-seed PMO replication (mean$\pm$standard deviation). With
$n = 3$, each two-sided 95\% $t$ interval is mean $\pm 2.484$ standard
deviations. C1 exceeds C4 on 8/8 tasks; the exact sign test gives one-sided
$p = .0039$ and two-sided $p = .0078$.}
\label{tab:seeds}
\end{table*}

We replicate C1 and C4 with three seeds on eight tasks spanning rediscovery,
activity, isomer, MPO, QED, and similarity objectives
(Table~\ref{tab:seeds}). C1 exceeds C4 on all eight tasks; the sum of
taskwise means is 5.55 versus 1.77, a 3.78-point (68\%) drop. An exact sign
test gives one-sided $p = 1/256 = .0039$ and two-sided $p = .0078$. The
smallest mean gap is 0.006 (QED), while the average gap is 0.473.
Retrieval-only uses a fixed seed and deterministic procedure, yielding one
score for each task.

The taskwise pattern separates three regimes. Direct reference access
saturates drd2, celecoxib rediscovery, and \texttt{isomers\_c7h8n2o2} in
every seed, producing large and stable C1--C4 gaps. Fexofenadine retains
useful blinded performance at $0.516 \pm 0.019$, alongside a 0.293 gap, while
task-agnostic QED changes by only 0.006. The larger C4 variance for drd2
shows that scalar-feedback search occasionally locates active regions, while
its mean remains 0.884 below C1. The multi-seed comparison therefore
reproduces both the aggregate reference effect and the objective-dependent
heterogeneity observed across all 23 tasks.

\subsection{Per-task heterogeneity}

The aggregate gap is concentrated but directionally uniform: C1 exceeds C4 on
all 23 tasks. The largest drops under blinding occur on drd2 (0.968),
\texttt{isomers\_c7h8n2o2} (0.961), troglitazone rediscovery (0.822),
mestranol similarity (0.816), celecoxib rediscovery (0.802), thiothixene
rediscovery (0.796), and jnk3 (0.719). The smallest gaps occur on
\texttt{valsartan\_smarts} (0.003), \texttt{amlodipine\_mpo} (0.037), and QED
(0.053). Table~\ref{tab:pertask} reports every condition for every task.

C4 retains useful feedback-driven search on individual tasks: it exceeds the
published REINVENT value on \texttt{albuterol\_similarity} (0.367 vs.\ 0.358)
and \texttt{ranolazine\_mpo} (0.358 vs.\ 0.297). Across all 23 tasks, the
aggregate C1 headline combines those feedback-driven gains with much larger
gains on identity-bearing tasks.

\subsection{Interpretation and a retrieval-free positive control}

\paragraph{Measured scope of C1--C5.}
For this representative agent and 50-call protocol, the fixed retrieval
policy attains 84\% of C1, while oracle-form information accounts for 0.54 of
the 9.59-point C1--C4 gap. This supports a retrieval-based explanation for
most of C1. Curated task-specific references also occur in ChatDrug
\citep{liu2024chatdrug}, AgentDrug \citep{le2025agentdrug}, and MT-Mol
\citep{kim2025mtmol}, motivating retrieval-safe ablations alongside their
standard settings.

\paragraph{Why reference exposure dominates.}
The mechanism is concrete on named-target tasks. For drd2, the Phase-0
dictionary supplies haloperidol, a known D2 antagonist whose direct oracle
score is approximately 0.99. Once that molecule is scored, the optimization
loop can obtain a high AUC by retaining it and proposing conservative
variants; discovering a D2-active scaffold from scalar feedback is no longer
necessary. In the unified all-task factorial, hiding names but retaining
references leaves drd2 and jnk3 at 0.980 and 0.745, whereas removing
references from the named prompt reduces them to 0.017 and 0.015
(Table~\ref{tab:pertask}). The pattern is therefore not confined to a small
diagnostic subset.

This distinction also changes how related systems should be evaluated.
Retrieval from a database supports applications that explicitly request known
scaffolds or analogues. For sample-efficient optimization claims,
component-aligned reporting supplies the LLM system and classical baseline
with matched task information and includes both standard and retrieval-closed
settings.

\paragraph{Positive control under strict blinding.}
We also evaluate \textsc{MolLEO-v4}, a population-based GA with an LLM
batch-mutation operator, structured memory, and adaptive planning under
strict blinding of task names, descriptions, and reference compounds. At the
matched 50-call AUC budget it reaches 7.39 versus REINVENT's 7.06. At 100
calls its sum best-score is $10.24 \pm 0.54$ (mean$\pm$standard deviation
over three seeds), compared with the published REINVENT sum-best of 8.72. In
a GPT-4o same-backbone comparison, it has higher per-task best scores than a
blinded SEISMO implementation on 16 of 23 tasks (aggregate 10.24 versus
9.24). The matched 50-call result measures sample efficiency, and the
100-call result measures achievable performance under blinding. Together with
the two favorable C4 comparisons above, this control provides a positive
characterization: after entity-level retrieval is closed, useful performance
remains in feedback-driven and population-level search. Component-level
ablations can further attribute the residual gain among chemical priors,
proposal diversity, memory, planning, and their interactions.

\paragraph{Recommended protocol.}
Report the standard prompt, C5-style identity-blinded oracle-form prompt, and
fully anonymous C4 prompt. The three-way comparison distinguishes direct
identity/reference access, oracle-form information, and scalar-feedback
optimization without changing the oracle.

\section{The Protein Story: Random-Walk Mutation as a Lower Bound}
\label{sec:protein}

PMO uses ML-based or formula-matching oracles whose names leak the answer.
The natural step toward ``real'' lead optimization is a docking-based oracle:
AutoDock Vina, GNINA, or a co-folding model such as Boltz-2
\citep{passaro2025boltz2}. Multiple recent agent papers report large gains in
this regime \citep{le2025agentdrug,kruger2026seismo}. We construct a 50-pair
protein-targeted lead-optimization benchmark with a multi-objective composite
oracle and measure agent gains relative to a matched random-walk floor.

\subsection{The \textsc{ProteinLeadOpt}-50 benchmark}
\label{sec:benchmark}

We select 10 medically-relevant protein targets from the \textsc{Dockstring}
dataset \citep{garciaortegon2022dockstring} spanning kinase (ABL1, JAK2,
MAPK14), GPCR (ADRB2, DRD2), nuclear-receptor (ESR2, PGR), DNA-repair
(PARP1), serine-protease (F2 thrombin), and CYP (CYP3A4) families. For each
target we sample 5 starter molecules with Vina docking scores in the 30--50\%
percentile of that target's \textsc{Dockstring} distribution, yielding
$10 \times 5 = 50$ (target, starter) pairs. Starters are reasonable but not
optimal hits, representing realistic lead-optimization headroom.

The composite oracle is
\begin{equation}
O(S) = I(S) + 0.5 \cdot \mathrm{QED}(S) - 2 \cdot P(S, S_0)
\end{equation}
with docking-improvement term
$I(S) = \min(1, \max(0, (d_{\mathrm{base}} - d(S))/5))$ where $d(S)$ is the
Vina docking score and $d_{\mathrm{base}}$ the starter's, drug-likeness
$\mathrm{QED}(S)$ \citep{bickerton2012quantifying}, and similarity penalty
$P(S, S_0) = \max(0, 0.3 - \tau(S, S_0))$ where $\tau$ is the
Morgan-fingerprint Tanimoto similarity to the starter
\citep{bajusz2015tanimoto}. The composite rewards docking improvement up to
5 kcal/mol, rewards drug-likeness, and heavily penalises divergence from the
starter ($\tau < 0.3$). Vina is evaluated via the public \textsc{Dockstring}
wrapper at ${\sim}6$\,s per molecule on 8 CPU cores. The oracle budget is 30
calls per pair.

We evaluate four methods: \textsc{LeadOpt-Agent} adapted to protein lead
optimization, with the top-five \textsc{Dockstring} molecules per target as
references; GPT-direct, one gpt-4o-mini prompt for 30 candidates; GraphGA,
with population size 20; and Random Walk, a fixed random RDKit single-edit
policy drawing from eight substituents (F, Cl, Br, CF3, OMe, Me, OH, NH2) at
aromatic H-bearing positions.

\subsection{Results and robustness}

\begin{table}[t]
\centering
\fontsize{7}{8.4}\selectfont
\setlength{\tabcolsep}{4pt}
\begin{tabular}{lcccc}
\toprule
Method & $\Delta$Dock & Succ. & Comp. & QED \\
       & (kcal/mol)   & rate  & mean  & mean \\
\midrule
GPT-direct    & $-0.36$          & 0.24 & 0.51              & 0.80 \\
LLM agent     & $-0.81$          & 0.36 & 0.55              & 0.78 \\
GraphGA       & $-0.84$          & 0.40 & 0.57              & 0.78 \\
RW (initial)  & $-0.98$          & 0.60 & 0.58              & 0.77 \\
RW (3 seeds)  & $-1.11 \pm 0.04$ & ---  & $0.636 \pm 0.008$ & ---  \\
\bottomrule
\end{tabular}
\caption{Aggregate \textsc{ProteinLeadOpt}-50 results at 30 calls over all 50
pairs. The first four rows give the matched single-seed comparison; the final
row gives the three-seed Random Walk replication (mean$\pm$standard
deviation). Success and QED were recorded for the matched run.}
\label{tab:protein}
\end{table}

In the matched run, Random Walk has the strongest docking improvement,
success rate, and composite score; its mean QED (0.77) is slightly below the
other methods (0.78--0.80). It wins the paired $\Delta$Dock comparison
against the LLM agent on 33/50 pairs. Three independent Random Walk seeds
over all 50 pairs yield composite $0.636 \pm 0.008$ and $\Delta$Dock
$-1.11 \pm 0.04$ kcal/mol (mean$\pm$standard deviation). Approximate 95\% $t$
intervals computed from the three rounded seed-level values are
$[0.611, 0.661]$ and $[-1.22, -1.00]$, respectively; the three seed-level
observations produce correspondingly wide intervals.

The operator-sensitivity analysis in Table~\ref{tab:sensitivity} changes both
substituents and eligible positions. Reducing eight substituents to F/Cl/Me
lowers composite from 0.637 to 0.510 and changes $\Delta$Dock from $-1.11$ to
$-0.55$. Allowing edits at any H-bearing atom yields 0.612 and $-0.99$. Thus
the broad substituent set matters, but the baseline remains competitive with
or stronger than the LLM agent (0.55 and $-0.81$) under the tested variants.

\paragraph{Why this happens.}
Adding a single Cl, Br, or CF3 at an aromatic position often lowers Vina
score by 0.5--1.5 kcal/mol through its volume and lipophilicity terms. Such
edits preserve the Murcko scaffold by construction, typically keep Tanimoto
similarity in the 0.85--0.95 range, and change QED by less than 0.1, so the
similarity penalty rarely activates. A concrete trajectory starts from the
JAK2 molecule \texttt{CN1CCC(=Cc2c[nH]c3ccc(F)cc23)CC1}, whose baseline
docking score is $-8.3$ kcal/mol. Random Walk adds one chlorine at an
aromatic position and reaches $\Delta$Dock $= -1.2$ kcal/mol within nine
calls. This chemically conservative, lead-preserving edit illustrates an
oracle landscape that systematically rewards accessible local lipophilic
changes.

Table~\ref{tab:pertarget} gives the per-target breakdown: Random Walk wins on
7 of 10 targets, GraphGA on ESR2 and CYP3A4, and the LLM agent on F2. Broader
graph mutations help GraphGA on ESR2 and CYP3A4, while the LLM agent performs
best on F2, where the kinase-style aromatic substitution pattern is less
reliably useful. Random Walk therefore provides a strong floor across most
targets and a target-level diagnostic for the remaining cases.

\paragraph{Implication for benchmark design.}
This result covers the AutoDock Vina + composite + 30-call combination. More
discriminative evaluations could use stronger oracles (FEP or Boltz-2
\citep{passaro2025boltz2}), selectivity objectives, and explicit novelty
requirements. Raising a lower-bound constraint from $\tau \geq 0.3$ to
$\tau \geq 0.5$ preserves leads more tightly while admitting the single
additions here, which typically have $\tau = 0.85$--$0.95$. A
scaffold-hopping benchmark needs an upper-bound similarity constraint
together with an explicit Murcko-scaffold change. Such a gate excludes the
local edits diagnosed here and creates room to test whether an agent can
propose genuinely novel structures. The random walk supplies a diagnostic
floor on the same oracle and budget.

\section{Discussion}
\label{sec:disc}

\paragraph{Scope of the findings.}
The experiments establish two scoped findings. For \textsc{LeadOpt-Agent} at
50 calls, direct reference submission reproduces most of the aggregate C1
score, while the retrieval-free positive control exceeds REINVENT at 50
calls. For \textsc{ProteinLeadOpt}-50, the random-walk advantage
characterizes the tested Vina-based oracle--budget combination. These
measurements separate application-useful scaffold retrieval, feedback-driven
optimization, and permissive local oracle directions.

\paragraph{What remains after retrieval is closed.}
The per-task C4 comparisons and the retrieval-free \textsc{MolLEO-v4} control
provide a constructive answer to the title question. Once entity-level keys
are removed, the positive evidence shifts from prompt-side recall toward
feedback-driven and population-level search: the blinded agent beats REINVENT
on two tasks, while the population agent exceeds it in aggregate at the
matched AUC budget. Component-matched ablations within a retrieval-closed
architecture can further attribute these gains among chemical priors,
proposal diversity, memory, planning, and their interactions.

\paragraph{Two distinct benchmark failure modes.}
The PMO and protein results diagnose different benchmark mechanisms. In PMO,
task identifiers and curated actives create an entity-retrieval route for the
agent; holding the oracle fixed while changing task-facing information
measures that route. In \textsc{ProteinLeadOpt}-50, every method receives the
same oracle and starters, and the oracle--constraint--budget combination
contains an easy local direction that a fixed random mutator exploits.
Retrieval-safe prompts measure the first mechanism, and an oracle-matched
search floor measures the second. Reporting both controls assigns observed
gains to their corresponding information and search channels.

\paragraph{Evidence beyond one prompt and backbone.}
A code audit finds direct target identifiers or target-specific retrieval
instructions in the task-described settings of public SEISMO
\citep{kruger2026seismo}, MolLEO \citep{wang2025molleo}, and AgentDrug
\citep{le2025agentdrug} implementations. In the inspected snapshots, SEISMO
names targets and supplies target/reference molecular information, MolLEO
maps drd2 to a dopamine type-2 receptor description, and AgentDrug asks the
model to search for an existing drug molecule for the target protein. These
observations document task-information exposure; SEISMO v3 additionally
reports a fully generic no-description condition.

Complementary trajectory probes with Claude Opus 4.5 show behavior consistent
with implicit retrieval. In one of five no-description drd2 runs, a
piperazine--benzimidazole D2 pharmacophore appears at step 3 and the score
jumps from 0.0002 to 0.64; the other trajectories lack this target-specific
jump. In the three-run jnk3 probe set, one trajectory explicitly proposes a
``kinase inhibitor-like scaffold'' although the target class is absent from
the prompt; analogous target-class scaffolds appear in the three-run gsk3b
probe set. The small probe set supplies exploratory evidence across a second
backbone. Together with the GPT-4o same-backbone comparison in
Section~\ref{sec:pmo}, it motivates full blinding evaluations with each
agent's intended backbone.

\paragraph{Cross-agent backbone control.}
We test prompt--backbone compatibility by running a SEISMO-style trajectory
prompt with gpt-4o-mini. On \texttt{perindopril\_mpo}, all 51 iterations emit
invalid variants of a ring-unclosed scaffold despite repeated parser
feedback; related mode collapse appears on three other tasks. We use this run
as a compatibility diagnostic and base the cross-agent performance comparison
on the later GPT-4o matched evaluation. A complete cross-agent study pairs
each agent's intended backbone with its standard and retrieval-closed
prompts.

\paragraph{Recommendations.}
We recommend two inexpensive checks. First, report C1, C5, and C4: this
distinguishes access to molecular identity, access to oracle-form
information, and scalar-feedback optimization. Running each condition with
the method's intended backbone preserves prompt--backbone compatibility.
Second, report a Random Walk floor on each docking-based composite oracle
using the same starters and budget. The baseline is short, deterministic, and
uses the same oracle-call budget as the agent. The agent--floor gap then
measures performance beyond fixed local search.

For the information control, authors should publish the exact field-level
prompt differences, keep the oracle, budget, validation, and generic probes
fixed, and report a retrieval-only floor when references are part of the
standard condition. For the oracle control, the baseline report should
specify its edit set and eligible positions and include docking change,
success rate, composite score, QED, similarity, and scaffold preservation.
These details attribute observed gains among search, initialization, and
objective geometry.

\section{Conclusion}

On PMO, a no-LLM retrieval baseline reproduces 84\% of
\textsc{LeadOpt-Agent}'s full-information score. Revealing only the oracle
form while hiding task/drug/target identity and reference structures improves
0.54 AUC over full blinding, far less than the 9.05-point loss from removing
identity and references. A complete 23-task factorial and an eight-task,
three-seed replication establish the pattern across tasks and seeds.
\textsc{MolLEO-v4} provides a positive example of optimization under strict
blinding at matched 50-call and extended 100-call budgets. On
\textsc{ProteinLeadOpt}-50, the three-seed Random Walk result and operator
sensitivities show that a Vina composite can reward simple local edits. These
findings motivate retrieval-safe prompts and oracle-matched trivial search
floors for low-budget benchmark claims.

The next measurement target is a low-budget benchmark with anonymous
task-facing fields, a more accurate affinity oracle such as FEP or a
validated co-folding model, selectivity against off-targets, and a novelty
rule that requires a scaffold change. This design measures transferable
structural reasoning under retrieval-safe information and a discriminative
molecular objective.

\section*{Limitations}

\paragraph{Stochastic coverage.}
The full 23-task factorial and C5 evaluation use one run per condition. The
core C1--C4 contrast uses three seeds on eight representative tasks, and seed
0 covers all 23 tasks. On \textsc{ProteinLeadOpt}-50, Random Walk uses three
seeds, while the matched LLM, GraphGA, and GPT-direct rows use one seed. The
reported $t$ intervals use three seed-level observations and are
correspondingly wide.

\paragraph{Agent and backbone scope.}
The causal C1--C5 ablations concern one gpt-4o-mini agent. Code inspection,
GPT-4o comparisons, and Opus trajectory probes extend the evidence across
architectures and backbones. The benchmark- and protocol-level
recommendations follow from the controlled \textsc{LeadOpt-Agent} ablations;
agent-specific estimates require the same ablation with each agent's intended
backbone.

\paragraph{Scope of C5.}
C5 preserves the objective family or metric and the optimization direction
while removing target-specific semantics. For celecoxib rediscovery, the LLM
sees ``ECFP4 Tanimoto similarity to a hidden reference. Maximize
similarity.'' C5 therefore measures oracle-form information after explicit
identity cues are removed. Future ablations can add graded, identity-free
task semantics as a separate factor.

\paragraph{Positive-control comparability.}
The 50-call \textsc{MolLEO-v4} comparison to REINVENT is metric- and
budget-matched and measures sample efficiency. Its 100-call sum-best result
and the 16/23 SEISMO comparison use twice SEISMO's 50-call budget and measure
achievable performance under blinding.

\paragraph{Protein benchmark scope.}
\textsc{ProteinLeadOpt}-50 uses one approximate oracle, one composite, and a
30-call budget. The random-walk conclusion may change with FEP, Boltz-2
\citep{passaro2025boltz2}, selectivity terms, or explicit scaffold-hopping
constraints. The sensitivity analysis also shows that the size of the
substituent set materially affects the baseline, even though the qualitative
diagnostic survives the tested variants.

\section*{Ethics Statement}

This work develops LLM agents for molecular optimization in drug discovery.
We acknowledge two ethical considerations.

\paragraph{Dual-use.}
Tools that improve binding to a protein target can in principle be applied to
design molecules with harmful effects, including toxic compounds or
controlled substances. We limit the experiments to public therapeutic
benchmark targets (kinases, GPCRs, and nuclear receptors), offline evaluation
with public docking software, and pretrained general-purpose models. The
study operates on the curated benchmark targets through fixed experimental
scripts.

\paragraph{Reproducibility and access.}
The experiments use PMO, TDC, RDKit, \textsc{Dockstring}, and AutoDock Vina.
We specify the model snapshot, prompts, oracle budgets, information
conditions, scoring rules, and optimization hyperparameters;
Appendix~\ref{sec:prompts} reports the prompt templates verbatim.

\paragraph{Computational cost.}
The complete PMO evaluation comprises 170 jobs and 8{,}400 oracle calls; its
2{,}128 reported API responses cost \$0.62 from recorded usage (\$0.63
including discarded smoke tests). Each protein optimization run uses roughly
$30 \times 6 \approx 180$ seconds of CPU-Vina docking and ${\sim}\$0.05$ in
API cost. We use pretrained API models without additional training.

\section*{Acknowledgments}

We thank the anonymous reviewers and area chairs for their constructive
feedback.

\bibliography{custom}

\appendix

\section{Prompt Templates}
\label{sec:prompts}

\subsection{Standard (C1) prompt for Phase 3}

\begin{verbatim}
You are an expert medicinal chemist
optimizing molecules for an oracle. Use
chemistry knowledge AND oracle feedback
to propose new SMILES that score higher.
Task: {TASK_NAME}
Reference compound: {REFERENCE}
Oracle hint: {ORACLE_DESCRIPTION}
Hypothesis: {HYPOTHESIS}
Top 5 trials so far: {TRAJECTORY}
Strategy: {EXPLOIT or DIVERSIFY}
Propose 5 new SMILES with rationales.
Return JSON: {"edits": [{"smiles":"...",
"rationale":"..."}, ...]}
\end{verbatim}

\subsection{Identity-Blinded Oracle-Form (C5) prompt}

C5 keeps the C1 prompt structure but replaces \texttt{TASK\_NAME} with an
anonymous identifier, sets \texttt{REFERENCE} to \texttt{None}, removes
task-specific seeds, and states only the objective family or metric and
optimization direction. The following is the actual celecoxib-rediscovery
prompt; no parenthetical target-class or structural description is supplied:

\begin{verbatim}
Task: task_42
Reference compound: None
Oracle hint: ECFP4 Tanimoto similarity
to a hidden reference. Maximize similarity.
\end{verbatim}

\subsection{Blinded (C4) prompt for Phase 3}

The differences from C1, applied uniformly:

\begin{itemize}
\item \texttt{TASK\_NAME} replaced by anonymous identifier
\texttt{task\_42}.
\item \texttt{REFERENCE} replaced by \texttt{None} (the dictionary is
empty).
\item \texttt{ORACLE\_DESCRIPTION} replaced by ``Unknown oracle. Probe to
learn its shape.''
\item Phase 1 probes use only the generic probes (benzene, biphenyl,
aspirin) since the reference dictionary is empty; the agent has no
task-specific seeds.
\end{itemize}

All other prompt machinery---system message, JSON schema, validation logic,
plateau detection, parallelism---is identical between conditions.

\section{Hyperparameters}

\paragraph{Phase 3 iterative loop.}
\begin{itemize}
\item LLM: \texttt{gpt-4o-mini-2024-07-18}, temperature 0.7, JSON response
format
\item Per-iteration proposals: $n = 5$
\item Plateau improvement threshold: 0.005
\item Phase 3 iterations: budget-limited; Phase 4 requests local analogues
\item Completion rule: deterministic valid, unique one-atom additions fill
any remaining budget, yielding exactly 50 calls in every LLM job
\item Oracle/deduplication: valid raw SMILES are scored; canonical SMILES are
used only for duplicate detection
\item Primary AUC: trapezoidal area over 50 running-best points divided by 50
(maximum 0.98)
\end{itemize}

\paragraph{Parallelism (\textsc{ProteinLeadOpt}-50).}
\begin{itemize}
\item Process workers per benchmark run: 15
\item Docking threads per worker: 6
\item Persistent dock cache: (target, canonical SMILES) $\rightarrow$ Vina
score, file-locked JSON
\end{itemize}

\section{Reference Compound Dictionary}

C1 and C3 use a fixed task-specific reference pool. It contains target
structures for rediscovery and similarity objectives, starting structures for
the two hopping objectives, known actives for the activity oracles, and
reference structures for the MPO objectives. C2, C4, and C5 use empty
reference and task-specific seed pools. Retrieval-only uses the same fixed
pool as C1 and C3, followed by deterministic one-atom variants.

\section{Additional Results}

The following tables provide the per-target protein breakdown, per-task PMO
results, multi-seed replication, and Random-Walk sensitivity referenced in
the main text.

\begin{table*}[t]
\centering
\small
\setlength{\tabcolsep}{3.5pt}
\begin{tabular}{lccccccc}
\toprule
Task & REINVENT & C1 & C2 & C3 & C4 & C5 & Retr. \\
\midrule
albuterol\_similarity        & .358 & .980 & .454 & .980 & .367 & .341 & .980 \\
amlodipine\_mpo              & .432 & .361 & .407 & .361 & .323 & .411 & .361 \\
celecoxib\_rediscovery       & .262 & .980 & .232 & .980 & .178 & .232 & .980 \\
deco\_hop                    & .558 & .659 & .515 & .789 & .517 & .518 & .490 \\
drd2                         & .195 & .980 & .017 & .980 & .012 & .012 & .980 \\
fexofenadine\_mpo            & .573 & .716 & .464 & .854 & .502 & .510 & .916 \\
gsk3b                        & .141 & .472 & .027 & .470 & .041 & .189 & .535 \\
isomers\_c7h8n2o2            & .078 & .980 & .226 & .980 & .019 & .010 & .980 \\
isomers\_c9h10n2o2pf2cl      & .257 & .133 & .112 & .008 & .028 & .001 & .000 \\
jnk3                         & .075 & .772 & .015 & .745 & .053 & .017 & .738 \\
median1                      & .139 & .392 & .086 & .392 & .102 & .072 & .392 \\
median2                      & .154 & .355 & .114 & .355 & .114 & .097 & .355 \\
mestranol\_similarity        & .275 & .980 & .144 & .980 & .164 & .181 & .980 \\
osimertinib\_mpo             & .670 & .517 & .625 & .728 & .391 & .635 & .350 \\
perindopril\_mpo             & .354 & .651 & .320 & .275 & .141 & .271 & .018 \\
qed                          & .905 & .882 & .868 & .780 & .829 & .837 & .000 \\
ranolazine\_mpo              & .297 & .504 & .031 & .358 & .358 & .293 & .112 \\
scaffold\_hop                & .422 & .708 & .376 & .653 & .361 & .379 & .857 \\
sitagliptin\_mpo             & .095 & .278 & .599 & .313 & .006 & .015 & .180 \\
thiothixene\_rediscovery     & .261 & .980 & .227 & .980 & .184 & .216 & .980 \\
troglitazone\_rediscovery    & .197 & .980 & .187 & .980 & .158 & .152 & .980 \\
valsartan\_smarts            & .000 & .003 & .000 & .001 & .000 & .000 & .000 \\
zaleplon\_mpo                & .364 & .177 & .023 & .041 & .000 & .000 & .000 \\
\midrule
Sum AUC & 7.06 & 14.44 & 6.07 & 13.98 & 4.85 & 5.39 & 12.16 \\
\bottomrule
\end{tabular}
\caption{Per-task top-1 AUC at 50 calls (seed 0). C1--C5 are defined in
Section~\ref{sec:pmo}; Retr.\ is retrieval-only. Sum values use unrounded
cells.}
\label{tab:pertask}
\end{table*}

\begin{table}[t]
\centering
\small
\begin{tabular}{lcccc}
\toprule
Target & LLM & GA & GPT & RW \\
\midrule
ABL1    & $-0.82$ & $-0.16$ & $+0.16$ & $\mathbf{-0.96}$ \\
ADRB2   & $-0.88$ & $-0.68$ & $-0.46$ & $\mathbf{-0.94}$ \\
DRD2    & $-0.86$ & $-0.86$ & $+0.66$ & $\mathbf{-1.08}$ \\
ESR2    & $-0.50$ & $\mathbf{-1.28}$ & $-0.26$ & $-0.96$ \\
F2      & $\mathbf{-1.08}$ & $-0.80$ & $+0.20$ & $-0.66$ \\
JAK2    & $-0.98$ & $-0.72$ & $-0.62$ & $\mathbf{-1.12}$ \\
MAPK14  & $-0.72$ & $-0.84$ & $-0.80$ & $\mathbf{-0.98}$ \\
PARP1   & $-0.82$ & $-0.78$ & $-0.68$ & $\mathbf{-1.06}$ \\
PGR     & $-0.84$ & $-0.96$ & $-1.04$ & $\mathbf{-1.10}$ \\
CYP3A4  & $-0.62$ & $\mathbf{-1.30}$ & $-0.76$ & $-0.98$ \\
\bottomrule
\end{tabular}
\caption{Per-target mean $\Delta$Dock (kcal/mol; more negative is better) in
the matched single-seed run. Each method evaluates five pairs per target.}
\label{tab:pertarget}
\end{table}

\begin{table}[t]
\centering
\fontsize{7}{8.4}\selectfont
\begin{tabular}{lrrrr}
\toprule
Scope & Jobs & Oracle & API & Cost \\
\midrule
Seed-0 C1--C5     & 115 & 5{,}750 & 1{,}666 & \$0.493 \\
Retrieval-only    & 23  & 1{,}050 & 0       & \$0.000 \\
Extra multi-seed  & 32  & 1{,}600 & 462     & \$0.130 \\
\midrule
Total             & 170 & 8{,}400 & 2{,}128 & \$0.623 \\
\bottomrule
\end{tabular}
\caption{Completed PMO evaluation jobs. All 147 LLM jobs use exactly 50
oracle calls. Costs use recorded token counts; discarded smoke/interrupted
responses add \$0.010 to the full ledger.}
\label{tab:cost}
\end{table}

\begin{table}[t]
\centering
\fontsize{7}{8.4}\selectfont
\setlength{\tabcolsep}{4pt}
\begin{tabular}{lcc}
\toprule
Condition & Composite & $\Delta$Dock \\
\midrule
A: 8 subs., aromatic-H          & 0.637 & $-1.11$ \\
B: F/Cl/Me, aromatic-H          & 0.510 & $-0.55$ \\
C: 8 subs., any H-bearing atom  & 0.612 & $-0.99$ \\
A, 3 seeds & $0.636 \pm 0.008$  & $-1.11 \pm 0.04$ \\
LLM agent (matched run)         & 0.55  & $-0.81$ \\
\bottomrule
\end{tabular}
\caption{Random-Walk operator sensitivity on \textsc{ProteinLeadOpt}-50.
Docking values are kcal/mol; more negative is better.}
\label{tab:sensitivity}
\end{table}

\paragraph{Random-Walk output properties.}
The three seed-level results are (composite, $\Delta$Dock):
$(0.625, -1.06)$, $(0.645, -1.15)$, and $(0.638, -1.12)$. Single additions
preserve the Murcko scaffold by construction (100\% for valid outputs),
typically produce Tanimoto similarities of 0.85--0.95 to the starter, and
change QED by less than 0.1; the matched-run mean QED is 0.77. These
properties explain why the similarity penalty seldom activates.

\end{document}